\documentclass[letterpaper]{article} % DO NOT CHANGE THIS
\usepackage{aaai25}  % DO NOT CHANGE THIS
\usepackage{times}  % DO NOT CHANGE THIS
\usepackage{helvet}  % DO NOT CHANGE THIS
\usepackage{courier}  % DO NOT CHANGE THIS
\usepackage[hyphens]{url}  % DO NOT CHANGE THIS
\usepackage{graphicx} % DO NOT CHANGE THIS
\usepackage{natbib}  % DO NOT CHANGE THIS AND DO NOT ADD ANY OPTIONS TO IT
\usepackage{caption} % DO NOT CHANGE THIS AND DO NOT ADD ANY OPTIONS TO IT
\usepackage{algorithm}
\usepackage{newfloat}
\usepackage{listings}
\DeclareCaptionStyle{ruled}{labelfont=normalfont,labelsep=colon,strut=off} % DO NOT CHANGE THIS
\floatstyle{ruled}
\newfloat{listing}{tb}{lst}{}
\floatname{listing}{Listing}
\usepackage{epigraph}
\usepackage{amsmath}
\usepackage{amssymb}
\usepackage{color, colortbl}
\definecolor{Gray}{gray}{0.9}
\usepackage{booktabs}
\usepackage{multirow}
\usepackage{textcomp}
\usepackage{enumitem}
\usepackage{tabularx}
\usepackage{algpseudocode}
\definecolor{lightblue}{RGB}{203,236,250}
\usepackage{lipsum}
\definecolor{codegreen}{rgb}{0,0.6,0}
\definecolor{codegray}{rgb}{0.5,0.5,0.5}
\definecolor{codepurple}{rgb}{0.58,0,0.82}
\definecolor{backcolour}{rgb}{0.95,0.95,0.92}

\title{ImagePiece: Content-aware Re-tokenization\\for Efficient Image Recognition}
\author{
    Seungdong Yoa\textsuperscript{\rm 1}, Seungjun Lee\textsuperscript{\rm 1}, Hyeseung Cho\textsuperscript{\rm 1}, Bumsoo Kim\textsuperscript{\rm 2 \dag}, Woohyung Lim\textsuperscript{\rm 1 \dag}
}
\affiliations{
    \textsuperscript{\rm 1}LG AI Research \textsuperscript{\rm 2}Chung-ang University\\
    \{seungdong.yoa, seungjun.lee, hs.cho, w.lim\}@lgresearch.ai, bumsoo@cau.ac.kr
}

\usepackage{bibentry}
\begin{document}

\maketitle

\begin{abstract}
    Vision Transformers (ViTs) have achieved remarkable success in various computer vision tasks.
    However, ViTs have a huge computational cost due to their inherent reliance on multi-head self-attention (MHSA), prompting efforts to accelerate ViTs for practical applications.
    To this end, recent works aim to reduce the number of tokens, mainly focusing on how to effectively prune or merge them.
    Nevertheless, since ViT tokens are generated from non-overlapping grid patches, they usually do not convey sufficient semantics, making it incompatible with efficient ViTs.
    To address this, we propose ImagePiece, a novel re-tokenization strategy for Vision Transformers.
    Following the MaxMatch strategy of NLP tokenization, ImagePiece groups semantically insufficient yet locally coherent tokens until they convey meaning.
    This simple retokenization is highly compatible with previous token reduction methods, being able to drastically narrow down relevant tokens, enhancing the inference speed of DeiT-S by 54\% (nearly 1.5$\times$ faster) while achieving a 0.39\% improvement in ImageNet classification accuracy.
    For hyper-speed inference scenarios (with 251\% acceleration), our approach surpasses other baselines by an accuracy over 8\%.
\end{abstract}
\section{Introduction}
\label{sec:intro}

\begin{figure}[t]
    \centering
    \includegraphics[width=\columnwidth]{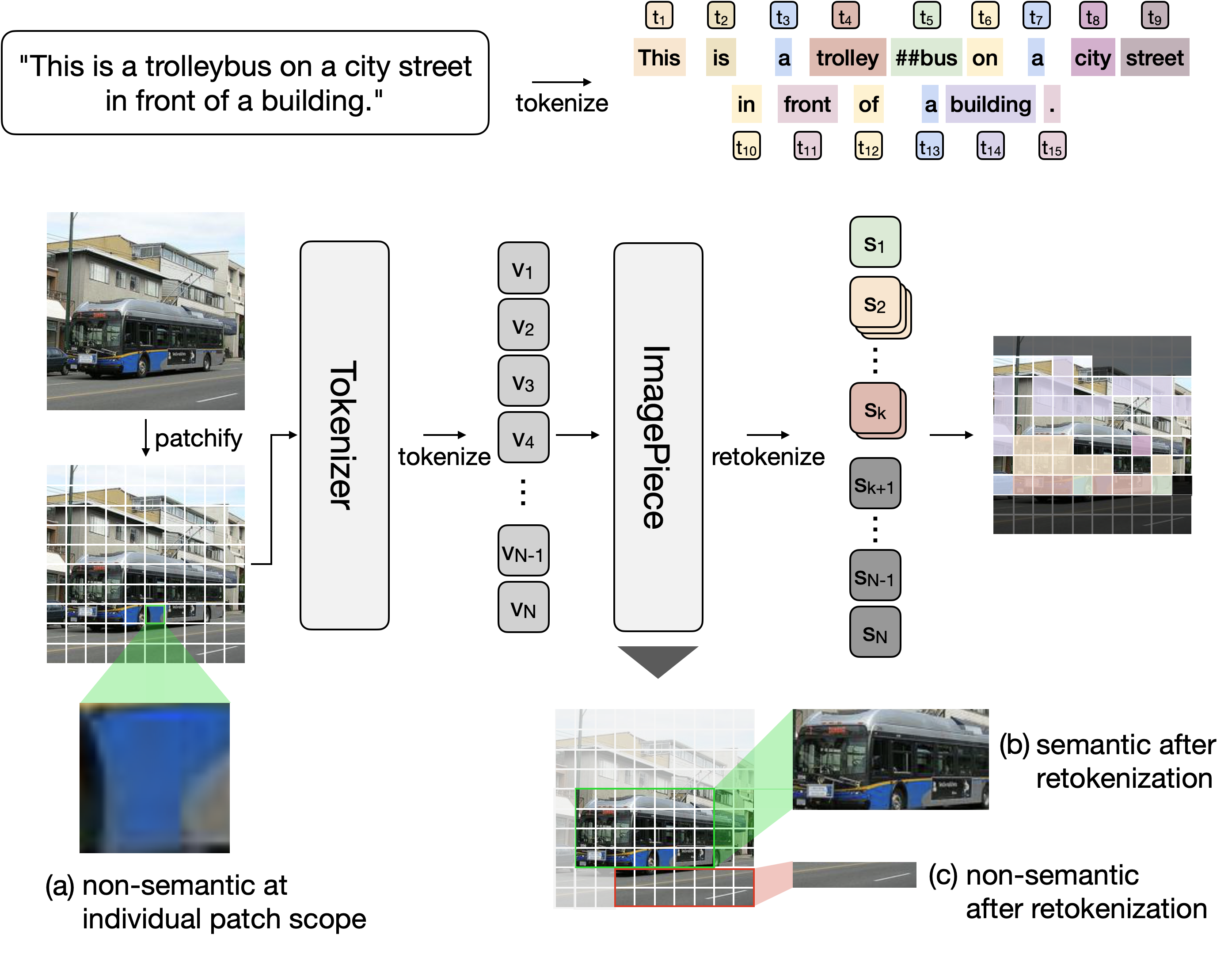}
    \caption{An illustration of the ImagePiece pipeline compared to WordPiece in NLP. While text is tokenized into meaningful tokens, image tokens from patches often contain irrelevant or non-semantic information. The retokenization in ImagePiece enables these non-semantic tokens to be merged into meaningful tokens, particularly when they have the potential to become meaningful after retokenization.}
    \label{fig:figure1}
\end{figure}

\setlength\epigraphwidth{\columnwidth}
\setlength\epigraphrule{0pt}
\epigraph{\textit{``All that is gold does not glitter, not all those who wander are lost."} --- J.R.R. Tolkien}

Tokenization involves breaking down text into smaller units called tokens, such as words, and is a crucial preprocessing step for nearly all Natural Language Processing (NLP) tasks.
Modern NLP models like BERT~\cite{devlin2018bert}, GPT~\cite{brown2020gpt}, and XLNet~\cite{yang2019xlnet} tokenize text into subword units.
These subword units strike a balance between words and characters, preserving linguistic meaning while minimizing out-of-vocabulary issues even with a relatively small vocabulary.
Subword tokenizer such as wordpiece~\cite{devlin2018bert} employs a greedy longest-match-first strategy, iteratively selecting the longest prefix of the remaining text that matches a vocabulary token, a method known as Maximum Matching or MaxMatch.

Following the pioneering success of Transformers in natural language processing, Vision Transformers~\cite{yin2022vit} have showcased remarkable performance across a broad spectrum of computer vision tasks.
In the context of images, tokenization involves splitting an image into non-overlapping grid patches and feeding the sequence of linear embeddings of these patches into a Transformer.
Here, image patches are treated similarly to tokens (words) in NLP.
However, unlike textual tokens stemmed from subwords where most elements contain meaning that contributes to understanding, image patch-based tokens differ significantly as they often 1) contain totally irrelevant semantics with the overall meaning (e.g., roads, sky, background) or 2) require a broader context to convey meaning (for example, in Fig~\ref{fig:figure1}, the blue patch itself (a) hardly contains any semantics while its collection with nearby patches (b) can be correctly recognized as a `bus').

This phenomenon becomes a more serious issue in further works streamlining ViTs; i.e., \textit{reducing} the number of input tokens~\cite{Liu_2021_ICCV,dong2022cswin,fan2021multiscale,li2022mvitv2,xiao2021early,chu2021twins,wang2021pyramid,graham2021levit}.
In ViTs, token reduction is mainly achieved through two primary methods: 1) \textit{token pruning}: removing inattentive tokens or 2) \textit{token merging}: fusing redundant (or similar) tokens into new abstractions.
Both approaches effectively reduce the complexity of Vision Transformers (ViTs).
However, due to the limitations of patch-based tokenization, this results in either 1) discarding tokens too hastily before their meaning has fully contextualized, or 2) prematurely smoothing semantic tokens with the nearest-similarity non-semantic noise, rather leading to more serious degradation than token pruning in certain scenes.

To address this, we propose \textbf{ImagePiece}, a novel re-tokenization strategy for Vision Transformers.
Following the MaxMatch strategy of NLP tokenization~\cite{devlin2018bert}, ImagePiece groups semantically insufficient (bottom-k) tokens to a level that convey meaning.
To add local inductive bias for visual scenes, we add a \textit{local coherence bias} module consisting of overlapping convolution layers to enhance non-semantic patches with similar positions to have higher similarity.
This facilitates the re-tokenization to merge nearby non-semantic tokens until they form meaningful tokens (or end up to be irrelevant with the overall meaning).
Then, the attention score of the semantic abstraction of inattentive tokens are measured again: if the abstraction gains relevance with the overall visual semantic, the token is re-organized.
For further efficiency, tokens that do not participate in the final semantic even after our re-tokenization are discarded.

ImagePiece poses several major advantages:
1) The locally coherent chunks of scattered inattentive tokens are facilitated to be merged to form a semantically meaningful abstraction, enabling the model to perform well-informed contextualization or pruning.
2) The attentive tokens that contain semantically important information are preserved from being merged (or smoothed), mitigating the limitation of previous token-merging pipelines.
Benefiting from these advantages, ImagePiece is able to drastically narrow down relevant tokens, enhancing the inference speed of DeiT-S by 54\% (nearly 1.5$\times$ faster) while achieving a 0.39\% improvement in ImageNet classification accuracy.
This is notable given that other baselines in efficient ViTs~\cite{liang2022evit,bolya2023token} suffer from an unignorable degradation in performance as a trade-off for this speed gain.
Our contributions are threefold:

\begin{itemize}
    \item We introduce ImagePiece, a novel retokenization strategy for efficient vision transformers that condense semantically irrelevant tokens. After condensing, the semantics of the tokens are re-evaluated, reorganizing tokens that have obtained meaning by gaining more contextual information.
    \item With local cohesive bias, non-semantic tokens that are positionally nearby are facilitated to be condensed together, adding local inductive bias of visual scenes. The condensed tokens that have no relevance after several iterations are discarded for efficiency.
    \item Our method achieves superior performance in ImageNet classification and demonstrates robust performance in hyper-speed inference, outperforming existing baselines.
\end{itemize}

\section{Related Work}
\label{sec:relatedWork}

\paragraph{Vision Transformers.}
Transformers~\cite{NIPS2017_3f5ee243}, originally prevalent in NLP, have recently attracted significant attention in computer vision, primarily due to their exceptional ability to model long-range dependencies.
Vision Transformers (ViTs)~\cite{dosovitskiy2020image} first introduced Transformer backbones to the field of computer vision, and a diverse array of studies~\cite{touvron2021training,yuan2021tokens,zhou2021deepvit,touvron2021going,pan2022less,pan2022fast,Liu_2021_ICCV,dong2022cswin,li2022mvitv2,xiao2021early,wang2021pyramid,graham2021levit} on ViTs have demonstrated their success in various aspects, ranging from architectural improvements to optimization techniques.

One of the core differences between NLP and computer vision applications of transformers lies in how input data is tokenized. In NLP, methods like WordPiece~\cite{devlin2018bert} and SentencePiece~\cite{kudo2018sentencepiece} tokenize text into subwords or characters, where each token carries meaningful semantic content. In contrast, when images are split into patches for ViTs, these patches are treated as tokens, yet each individual patch often lacks inherent semantic meaning. This fundamental difference presents unique challenges and opportunities in how transformers are applied across these domains.
\\

\paragraph{Efficient Transformers.}
Recent advancements in both NLP and computer vision domains have experienced a surge in efforts to enhance transformer models' efficiency.
These efforts include developing efficient attention mechanisms~\cite{kitaev2020reformer,bolya2022hydra,wang2020linformer,dao2022flashattention,shen2021efficient,choromanski2020rethinking}, pruning of transformer heads or features~\cite{michel2019sixteen,voita2019analyzing,meng2022adavit}, and integrating vision-specific modules~\cite{Liu_2021_ICCV,dong2022cswin,graham2021levit,mehta2021mobilevit}.
Several recent approaches in NLP~\cite{lassance2021study,kim2022learned,kim2020length,goyal2020power} and computer vision~\cite{meng2022adavit,ryoo2021tokenlearner,yin2022vit,kong2021spvit,song2022cp,fayyaz2022adaptive,yu2023unified,marin2021token,xu2022evo,pan2021ia,long2023beyond,liang2022evit,rao2021dynamicvit,bolya2023token,chen2023cf} have attempted to reduce the number of tokens due to the input-agnostic nature of transformers.
Especially, for ViTs, recent works to reduce the number of tokens have emerged by diverse approaches, such as token pruning and token merging.
In token pruning, DynamicViT~\cite{rao2021dynamicvit} introduces a method to prune tokens for a fully pre-trained ViT using additional training parameters.
EViT~\cite{liang2022evit} determines inattentive tokens according to class token attention, and discards these tokens to reorganize image tokens.
Token merging approaches, such as ToMe~\cite{bolya2023token}, combine the similar token pairs into new tokens to reduce the number of tokens.
Our approach provides a novel perspective for efficient ViTs by proposing the retokenization to initially merge non-semantic tokens into semantically meaningful chunks to correctly measure their relevance with the visual scenes.

\section{Preliminary}
\label{sec:pre}
In this preliminary section, we start with a basic overview of Vision Transformers (ViTs), i.e., how images are patchified and tokenized for Transformer architectures.
Next, we explain the basic concepts involved in evaluating each patch token.
Finally, we describe the challenges faced by previous token reduction methods.

\paragraph{ViT Overview.}
ViT first divides an input image into \textit{non-overlapping} $p \mkern-4mu\times\mkern-4mu p$ patches and projects each patch to a token embedding.
Typically, with a patch size of $16 \mkern-4mu\times\mkern-4mu 16$ ($p$=16) and an image size of $224 \mkern-4mu\times\mkern-4mu 224$, we obtain 196 image tokens.
An extra class token, denoted as \texttt{[CLS]}, is appended to the sequence of image tokens to serve as an aggregator of global image information for the final classification.
After all of the tokens are combined with positional embeddings, the patch embeddings are fed into a transformer encoder.

\paragraph{Evaluating Token Importance.}
Following previous work in literature~\cite{liang2022evit}, we define the class attention score as the measure of interaction between the class token and the image tokens, indicating the importance of each token in contributing to the overall semantics of an image.
Let $D$, $N$ denote the length of the query vector and the number of input tokens, respectively, where the input tokens in each transformer block refer to the class token and the remaining image tokens.
The token sequence $X$ is projected into a query matrix $\mathbf{Q} \in \mathbb{R}^{N \times D}$, a key matrix $\mathbf{K} \in \mathbb{R}^{N \times D}$, and a value matrix $\mathbf{V} \in \mathbb{R}^{N \times D}$.
The class attention score and the output of the class token are as follows:

\begin{equation}
x_{\text{class}} = \text{Softmax}\left( \frac{\mathbf{Q}_{\text{class}} \cdot \mathbf{K}^\top}{\sqrt{D}} \right) \mathbf{V} = A_{\text{class}} \cdot \mathbf{V},
\label{eq:attentive_score}
\end{equation}
where $\mathbf{Q}_{\text{class}}$ denotes the query vector of the class token.
As a result, the output of the class token, denoted as $x_\text{class}$, is a linear combination of the value vectors $\mathbf{V}$=$[v_1, v_2,\cdots,v_N]^\top$.
The coefficients of this combination, denoted by $A_{class}$ in Eq.(\ref{eq:attentive_score}), are the attention values from the class token with respect to all tokens.

\paragraph{Challenges in Token Reduction.}
\label{subsec:challenges}
Token reduction techniques, including token pruning and token merging, are essential for accelerating ViTs by decreasing the number of tokens.
Token pruning~\cite{liang2022evit,rao2021dynamicvit,kong2021spvit,yin2022vit,meng2022adavit} and token merging~\cite{bolya2023token,ryoo2021tokenlearner,marin2021token} are the two most representative branches of ViT acceleration.
Token pruning methods eliminate less attentive tokens based on a fixed hyper-parameter, targeting the fixed bottom-k tokens deemed unimportant or those under a predetermined threshold, as discussed in~\cite{liang2022evit,meng2022adavit}.
However, since tokens are either treated as important or non-important according to the attentiveness score with the \texttt{[CLS]} token, locally coherent tokens that might entail semantically crucial information are often discarded.
To compensate the information loss that often occurs in token pruning, token merging approaches attempted to accelerate ViTs by merging pairs of tokens with the highest similarity, selecting the top-k most similar pairs rather than removing original tokens.
However, as the merging process progresses, tokens with low similarity are prone to be merged, resulting in semantically essential tokens to be diluted to a semantically irrelevant interpolation of two unrelated representations.

\section{ImagePiece}
\label{sec:method}

\begin{figure*}[t]
    \centering
    \includegraphics[width=\textwidth]{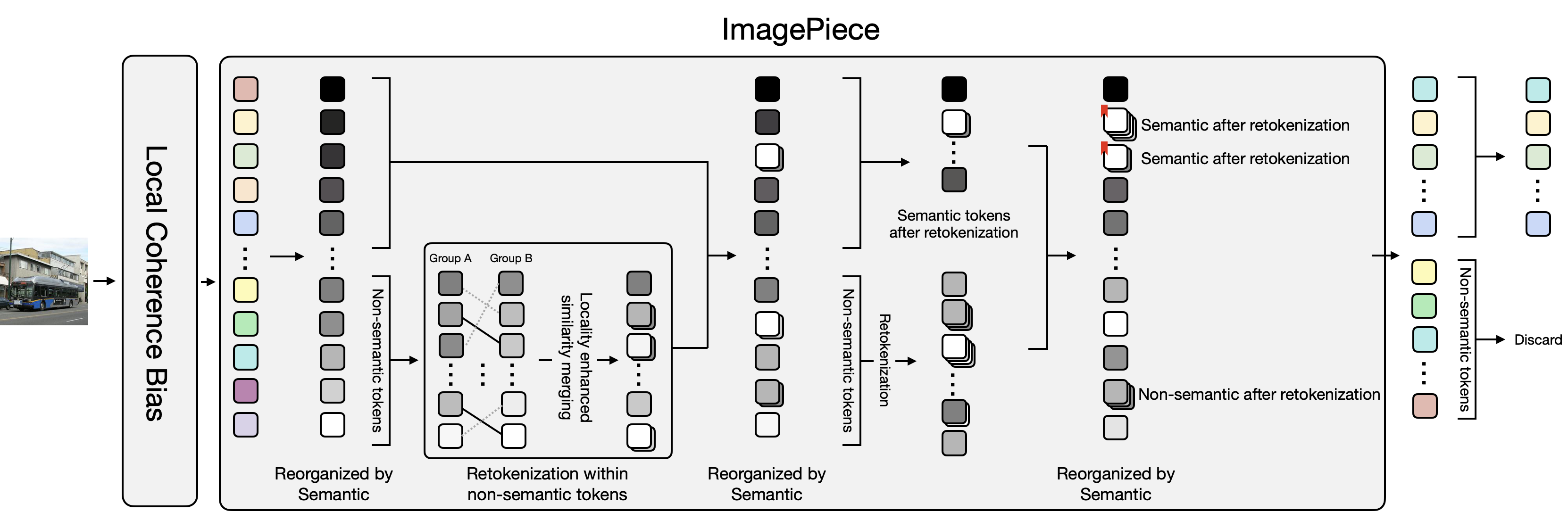}
    \caption{Overall architecture of the proposed method.}
    \label{fig:overall_architecture}
\end{figure*}

We introduce ImagePiece, a novel re-tokenization strategy that makes image patch tokens analogous to subwords (minimal unit that conveys meaning).
Fig.~\ref{fig:overall_architecture} illustrates the overall architecture of our proposed ImagePiece.

\paragraph{Re-tokenizing Non-semantic Tokens.}
\label{subsec:method_inattentive_token_merging}

Following the MaxMatch strategy of WordPiece tokenization~\cite{devlin2018bert}, ImagePiece performs retokenization for ViTs that groups semantically insufficient (bottom-k) tokens to a level that conveys meaning.
In detail, the re-tokenization iteratively follows a three-stage procedure.
Step I, the importance of each token to the overall semantic is evaluated by the attention score with the \texttt{[CLS]} token~\cite{liang2022evit} as in Eq.(\ref{eq:attentive_score}).
Step II, the bottom-$k$ tokens are divided into two equal-sized subsets, where the tokens organized by their attention are alternatively assigned to each group (for convenience, we refer to these groups as $A$ and $B$).
Then, each token in group $A$ is merged with the most similar token in group $B$ using bipartite soft matching~\cite{bolya2023token}.
Step III, the attention scores of the semantic abstractions of the matched tokens along with remaining non-bottom-$k$ tokens are recalculated: the abstraction tokens that gain relevance with the overall visual semantic are re-organized.

\paragraph{Local Coherence Bias.}

As illustrated in Fig.~\ref{fig:figure1}, patches whose meaning is hard to discriminate at an individual scope can gather meaningful semantics when contextualized with nearby patches (i.e., local inductive bias).
To add this bias to visual tokens, we add a \textit{local coherence bias} module consisting of overlapping convolution layers.
As adjacent patches are entangled with overlapping features, non-semantic patches that are geometrically close are encouraged to have higher similarity.
This facilitates the re-tokenization to merge nearby non-semantic tokens until they form meaningful tokens (or end up being irrelevant with the overall meaning).

\paragraph{Compatibility with Token-Pruning.}

As ViTs suffer from the quadratic complexity of input tokens and the square-grid patch tokenizer~\cite{yin2022vit} usually contains high portions of irrelevant or redundant patches, previous works have devised pruning strategies to \textit{drop-out} tokens that do not contribute to the overall semantic of a visual scene~\cite{rao2021dynamicvit,liang2022evit}.
However, patches whose semantic is difficult to determine without proper contextualization are prone to be discarded at earlier layers due to the progressively pruning strategy where the tokens are discarded at intermediate ViT layers.
On the other hand, due to this progressive architecture, ImagePiece shows powerful compatibility with existing token pruning methods, since the non-semantic tokens are quickly merged with geometrically nearby tokens, benefiting from strong local bias to form meanings at earlier transformer layers (see Fig.~\ref{fig:speed_and_acc}).

\paragraph{Competitiveness with Token-Merging.}
ImagePiece can be treated as a variation of token merging~\cite{bolya2023token,ryoo2021tokenlearner,marin2021token}.
Previous work in token merging is mainly focused on combining redundant tokens (with high feature similarities) to enhance efficiency in Vision Transformers.
However, this strategy entails serious drawbacks.
For instance, when the similarity between the top-$k$ similar token pairs are not sufficient enough, the merging pipelines above results in \textit{smoothing} highly semantic tokens with rather irrelevant tokens.
This leads to rather harming crucial semantics of the visual scene.
Tab.\ref{tab:merged_topk} shows empirical support that semantically-relevant tokens are frequently merged (thus being semantically diluted) in the initial ViT layers with previous token merging methods.
Conversely, ImagePiece retokenizes non-semantic tokens until they contribute to a meaningful semantic representation.
Experimental results show that due to this strategical difference, ImagePiece successfully preserves semantically essential tokens without interfering with highly-semantic tokens.

\section{Experiments}
\label{sec:experiments}
In this section, we conduct extensive experiments to demonstrate our retokenization strategy for efficient ViTs.
We first show that our method outperforms the baselines by a substantial margin.
Analytical experiments are conducted to further explain the factors contributing to our method's effectiveness.
Notably, the hyper-speed inference experiment in Fig.~\ref{fig:speed_and_acc} reveals that our method achieves relatively robust performance even with a drastically reduced number of tokens.\\

\paragraph{Implementation details.}

We conduct all the experiments on ImageNet-1k~\cite{deng2009imagenet} consisting of 1.2 million images in the training set and 50k images in the test set.
The image resolution is $224 \mkern-4mu\times\mkern-4mu 224$ in training and testing.
During training our models, we simply follow all the training strategies and optimization methods used in DeiT~\cite{touvron2021training}.
We train our model from scratch for 300 epochs, and we don't use any tricks (e.g., adding extra parameters, starting from an existing checkpoint or fine-tuning, using additional training tricks), unlike other prior works.
The throughput (img/s) is measured on a single NVIDIA GeForce RTX 3090 during inference.
% Also, we measure the multiply-accumulate computations (MACs) following~\cite{liang2022evit}.

For our method to apply the local coherence bias module, we adopt simple convolutional layers (four $3 \mkern-4mu\times\mkern-4mu 3$ convolutions and a single $1 \mkern-4mu\times\mkern-4mu 1$ convolution), replacing a standard ViT's patchify stem.
In all experiments, we set the proportion $p$ of the non-semantic token set to 0.3, targeting the bottom 30\% of tokens based on their importance (attentiveness).
Therefore, these tokens are candidates for retokenization.
We set the similarity merging ratio to 0.08, meaning that token pairs equivalent to 0.08 $\times$ the total number of tokens from the non-semantic token set are selected for merging based on their similarity.
Also, we set the pruning ratio $r$ to 0.8, which results in discarding the bottom 20\% of tokens, identified as non-semantic, after retokenization by ImagePiece.
These hyperparameters are specifically adjusted to further accelerate ViTs beyond the standard settings.

\renewcommand{\arraystretch}{1.25}
\begin{table}[t!]
    \centering
    \LARGE
    \resizebox{\columnwidth}{!}{%
        \begin{tabular}{l|cc}
        \multirow{2}{*}{Model} & Acc  & Throughput \\
        & (\%) & (img/s) \\        
        % Model & Acc (\%) & Throughput (img/s) & MACs (G) & Param (M) \\
        \hline
        DeiT-Ti & 72.13 & 6429.2 \\
        Pruning by learned projection layer~\cite{rao2021dynamicvit}  & 71.20 \color{blue}(-0.93) & 9351.0 \\  
        Pruning by learned token selector~\cite{kong2021spvit} & 72.09 \color{blue}(-0.04) & 9245.2 \\
        Pruning by [CLS] attentiveness~\cite{liang2022evit} & 71.70 \color{blue}(-0.43) & 9432.9 \\
        \rowcolor{lightblue}
        Pruning by retokenization (Ours) & \textbf{72.61} \color{red}(+0.48) & \textbf{9450.2} \\
        \hline
        \hline
        DeiT-S & 79.83 & 2531.1 \\
        Pruning by learned projection layer~\cite{rao2021dynamicvit} & 79.32 \color{blue}(-0.51) & 3762.0 \\
        Pruning by learned token selector~\cite{kong2021spvit} & 79.34 \color{blue}(-0.39) & 3722.1 \\
        Pruning by [CLS] attentiveness~\cite{liang2022evit} & 79.37 \color{blue}(-0.46) & 3787.5 \\
        \rowcolor{lightblue}
        Pruning by retokenization (Ours) & \textbf{80.22} \color{red}(+0.39) & \textbf{3891.9} \\
        \hline
        \end{tabular}%
    }
    \caption{Comparison to existing token pruning models with DeiT on ImageNet-1k. The color of the number indicates the performance gap compared to the original models, DeiT-Ti and DeiT-S. Ours outperforms all the baselines, especially the original DeiT-S improving throughput (img/s) by 54\%.}
    \label{tab:main_result}
\end{table}

\subsection{Main Results}
In Tab.~\ref{tab:main_result} and Tab.~\ref{tab:main_result_merge}, we describe the main results on ImageNet~\cite{deng2009imagenet} using the base models, DeiT-Ti and DeiT-S.
We report the top-1 accuracy (\%) and throughput (img/s).

\paragraph{Comparisons with prior pruning methods.}
To demonstrate the effectiveness of our retokenization approach in pruning, we compare our ImagePiece with prior pruning methods.
For token pruning, DynamicViT~\cite{rao2021dynamicvit} employs a \textit{learned projection layer}, SPViT~\cite{kong2021spvit} utilizes a \textit{learned token selector}, and EViT~\cite{liang2022evit} leverages a \textit{[CLS] attentiveness score}.
Overall, our method achieves the best accuracy with slightly faster inference speed as shown in Tab.~\ref{tab:main_result}.
Especially, compared to the original DeiT-Ti and DeiT-S, our method accelerates the inference speed by 47\% and 54\% respectively, while also notably improving the performance with an accuracy gap of 0.48\% for DeiT-Ti and 0.39\% for DeiT-S.

\paragraph{Comparisons to token merging models.}

As shown in Tab.~\ref{tab:merged_topk}, our observations reveal that important tokens are frequently merged in the initial layers of previous token merging method.
This indicates that important tokens are quickly merged in the early layers, and as the merging process progresses to later layers, the proportion of inattentive tokens increases, leading to lower similarity in the merging process at these stages.
Additionally, since merging tokens can potentially condense the information of target tokens, it may also reduce the granularity of details, particularly in important tokens, leading to a loss of critical information.
To address these issues, our method, ImagePiece, selectively merges only the less significant tokens.

In Tab.~\ref{tab:main_result_merge}, we compare our method with existing token merging methods: ToMe~\cite{bolya2023token} which uses bipartite soft matching algorithm for merging, Token Pooling~\cite{marin2021token} which is similar to ToMe~\cite{bolya2023token} but utilizes a slow K-Means based approach, and Token Learner~\cite{ryoo2021tokenlearner} which uses an MLP to reduce the number of tokens.
Unlike these methods, our approach focuses on transforming non-semantic tokens into semantically meaningful ones.
This crucial step in our retokenization process, implemented through ImagePiece, provides a significant advantage over conventional token merging strategies.
Tab.~\ref{tab:main_result_merge} shows that even without pruning, ImagePiece outperforms these baseline methods with faster inference speed.
Our method, which employs a pruning process after the retokenization of ImagePiece, achieves the best accuracy, outperforming existing token merging methods by 3.65\% on average.

\renewcommand{\arraystretch}{1.25} % 1.5 times the default row height
\begin{table}[t!]
    \centering
    \resizebox{\columnwidth}{!}{%
        \begin{tabular}{l|cc}
        \multirow{2}{*}{Model} & Acc  & Throughput \\
        & (\%) & (img/s)  \\
        \hline
        Token Pooling~\cite{marin2021token} & 71.35 \color{blue}(-8.48) & 3571.0 \\
        Token Learner~\cite{ryoo2021tokenlearner} & 79.01 \color{blue}(-0.82) & 3747.7 \\
        ToMe~\cite{bolya2023token} & 79.36 \color{blue}(-0.47) & 3806.1 \\
        \hline
        ImagePiece & 79.67 \color{blue}(-0.16) & 3890.1 \\
        \rowcolor{lightblue}
        ImagePiece + Pruning (Ours) & \textbf{80.22} \color{red}(+0.39) & \textbf{3891.9} \\
        \hline
        \end{tabular}%
    }
    \caption{Comparison to prior token merging methods with DeiT-S on ImageNet-1k.
    Ours outperforms all the baseline models by an average of 3.65\%.}
    \label{tab:main_result_merge}
\end{table}

\subsection{Extensive Experimental Results}
We also conduct extensive experiments, such as hyper-speed evaluation and random masking noise evaluation, to demonstrate the efficiency and effectiveness of our approach.
All experiments are conducted on 50k images from the ImageNet test set, and unless otherwise stated, the inference speed for all models is maintained at the same level.

\paragraph{Hyper-speed Inference Results.}
\label{sec:hyperspeed}
We conduct a hyper-speed evaluation experiment to demonstrate the effectiveness of our method in Fig.~\ref{fig:speed_and_acc}.
In this experiment, we use the models trained in typical settings mentioned in Tab.~\ref{tab:main_result} and Tab.~\ref{tab:main_result_merge}, e.g., DynamicViT with a keep rate of 0.7, EViT with a keep rate of 0.7, ToMe with a reduction factor $r$=13 (the number of merging tokens per layer), and ours with a keep rate of 0.8 and a merging rate of 0.08.
Using these models, we evaluate their performance at various higher speeds without further training.
For example, EViT initially trained at a keep rate of 0.7 is assessed at lower keep rates such as 0.6, 0.5, $\dots$, 0.27, and ToMe initially trained at $r$=13 is evaluated at higher $r$ values (e.g., 16, 18, $\dots$, 25) to further accelerate the inference speed.
The details of each model's setting are in the supplements.

The performance gap between the baselines and ours is gradually increasing (0.86\% to 8.15\% on average) as tokens are further reduced to accelerate the inference speed.
Specifically, when compared to DeiT-S, at a 2.51$\times$ speedup, the accuracy of DynamicViT, EViT, and ToMe decreases by 20.05\%, 13.99\%, and 9.24\% respectively, whereas our method shows only a 6.28\% decrease.
This highlights the superior efficiency of our approach in relatively maintaining accuracy under significant token reduction.
Additionally, Tab.~\ref{tab:hyperspeed_remainToken} demonstrates that our method achieves 79.38\% accuracy with only 26 output tokens for prediction, which represents just 13\% of the original total of 197 tokens (including \texttt{[CLS]}), compared to 68 (35\%), 69 (35\%), and 41 (21\%) output tokens for DynamicViT, EViT, and ToMe respectively.
This indicates that at a similar performance level, our method improves speed by approximately 30\% over the baselines with fewer tokens.
From the experiments, we demonstrate that our method successfully retokenizes semantically insufficient tokens into more meaningful units and effectively discards non-semantic tokens by more accurate re-evaluation.
This enables it to maintain relatively robust performance, even under drastic token reduction for faster inference speed.

{
\setlength{\abovecaptionskip}{0pt}
\begin{figure}[t!]
    \centering
    \includegraphics[width=0.8\columnwidth]{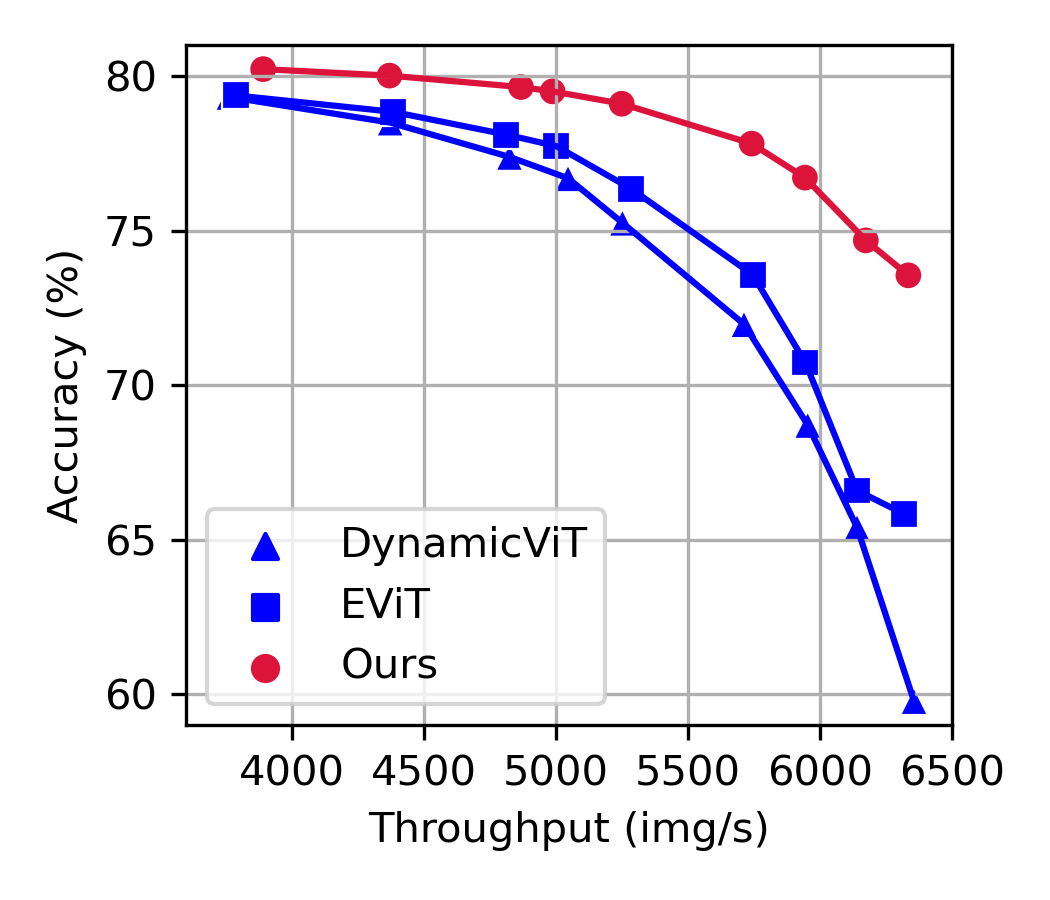}
    \caption{Comparison of our ImagePiece with the patch tokenizer of ViT in terms of inference speed and accuracy. While the baselines show a marked decline in performance as throughput increases, our method not only maintains relatively robust accuracy but also demonstrates compatibility, even at high speeds.
    }
    \label{fig:speed_and_acc}
\end{figure}
}

\paragraph{Model Robustness to Random Masking Noise.}

We also conduct a random masking noise experiment to further investigate the effectiveness of our retokenization method.
In Tab.~\ref{tab:random_masking_noise}, we report the performance of each method on randomly masked test samples.
16$\times$16 masks are applied to the test samples at random locations, and the evaluation is performed as the number of masks gradually increases.
In this experiment, we use the models in Tab.~\ref{tab:main_result} and Tab.~\ref{tab:main_result_merge}, based on DeiT-S, conducting evaluations through inference.

\noindent When the number of random masks is 7, only our method maintains performance over 79.5\%, with a performance gap of 1.4\% between ours and the baselines.
As the number of masks increases, this gap also increases; notably, at 50 masks, the performance gap between our method and the baselines averages 5.2\%.
This suggests that semantically meaningful tokens by retokenization are more helpful in comprehending a global visual scene with masking noise than individual patch-scope tokens, as it can achieve better performance than the baselines even with just a few non-masked areas of the image. Additionally, while the baselines exhibit a performance of around 79.3\% without noise in Tab.~\ref{tab:main_result} and Tab.~\ref{tab:main_result_merge}, our method surpasses this, reaching 79.63\%, even with 7 random masks present.

\begin{table}[t!]
    \centering
    \small
    \resizebox{\columnwidth}{!}{%
        \begin{tabular}{l|ccc}
        Model & Acc & Throughput & \#output tokens (\% of Total) \\
        \hline
        DynamicViT & 79.32 & 3762.0 & 68 (35\%)  \\        
        EViT       & 79.37 & 3787.5 & 69 (35\%) \\
        ToMe       & 79.36 & 3806.1 & 41 (21\%) \\
        \rowcolor{lightblue}
        Ours       & \textbf{79.38} & \textbf{4868.0} & \textbf{26} (13\%) \\
        \hline
        \end{tabular}%
    }

\caption{Comparison of throughput (img/s) and the number of output tokens after token reduction at a similar performance level.
Our method achieves approximately 30\% speed increase while retaining accuracy comparable to the baselines, with only 26 output tokens remaining, representing 13\% of the original total tokens.
}
\label{tab:hyperspeed_remainToken}
\end{table}

\begin{table}[t!]
    \centering
    \resizebox{\columnwidth}{!}{%
        \begin{tabular}{l|cccccc} 
        \multirow{2}{*}{Model} & \multicolumn{6}{c}{\# of Random Masks} \\
        & 7 & 10 &  15 & 20 & 25 & 50 \\
        \hline
        DeiT-S     & 79.05 & 78.65  & 77.78 & 76.52 & 75.29 & 68.28     \\        
        DynamicViT & 78.54 & 78.06  & 77.09 & 75.95 & 74.81 & 68.17     \\        
        EViT       & 78.21 & 77.72  & 76.77 & 75.42 & 74.47 & 68.10     \\
        ToMe       & 78.41 & 77.73  & 76.49 & 74.99 & 73.68 & 65.41     \\
        \rowcolor{lightblue}
        Ours       & \textbf{79.63} & \textbf{79.13} &  \textbf{78.36} & \textbf{77.20} & \textbf{76.63} & \textbf{71.42} \\        
        \hline
        \end{tabular}
    }

\caption{Random masking noise evaluation: comparison to baseline models on randomly masked test samples.
}
\label{tab:random_masking_noise}
\end{table}

\subsection{Analysis}
% We also present some interesting observations from the following experiments, including an analysis of retokenization, similarity merging, compatibility and ablation study.
% In this section, we present some interesting observations from analysis of ImagePiece.

\paragraph{Token Attentiveness after Re-tokenization.}
\label{subsec:supple_individual_vs_chunk}
We propose ImagePiece, which is a novel retokenization strategy for token reduction.
ImagePiece initially merges non-semantic tokens into semantically meaningful chunks to accurately assess their relevance within the global scene of the image.
After merging the tokens from the inattentive token set in a previous layer, we then discard the tokens considered inattentive by evaluating the attentiveness of all tokens.
Although some tokens are identified as inattentive in the previous layer, they can be reassessed as attentive tokens after the retokenization if they embody more semantic information as chunks.
Tab.~\ref{tab:inattn_to_attn} demonstrates how many tokens initially treated as inattentive can be reassessed as attentive when merged into chunks.
The ratio $\text{Ratio}_{\text{inattn}\to\text{attn}}$ in Tab.~\ref{tab:inattn_to_attn} represents the proportion of tokens that were initially deemed inattentive and subsequently retokenized in the previous layer, but are now evaluated as attentive in the current layer.
For example, 27.99\% of tokens deemed inattentive and retokenized in the 1-st layer are reassessed as attentive in the 2-nd layer, highlighting the capability of ImagePiece to reevaluate token significance.
This reevaluation is particularly pronounced in the 6-th layer, where 58.77\% of tokens that were considered inattentive and merged in the 5-th layer are reassessed as attentive.

\begin{table}[t!]
    \centering
    \Large
    \resizebox{\columnwidth}{!}{%
    \begin{tabular}{l|cccccc}
        Layer         & 2 & 3 & 5 & 6 & 8 & 9 \\
        \hline
        $\text{Ratio}_{\text{inattn}\to \text{attn}}$ (\%) & 27.99 & 44.39 & 35.49 & 58.77 & 28.97 & 43.27 \\
        \hline
    \end{tabular}
    }
    \caption{The percentage of previously inattentive tokens reassessed as attentive after retokenization.
    }
    \label{tab:inattn_to_attn}
\end{table}

\begin{table}[t]
    \centering
    \resizebox{\columnwidth}{!}{%
        \begin{tabular}{l|cc|c}
        Model         & First layer & Last layer & Gap \\
        \hline
        ToMe~\cite{bolya2023token}  & 0.7852 & 0.6763 & \color{blue}{-0.1089} \\
        ImagePiece (Ours) & 0.8091 & 0.8096 & \color{red}{+0.0005}\\
        \hline
        \end{tabular}%
    }
\caption{Comparison of similarity scores among merged token pairs for each sample on the ImageNet test set.
We calculate the average of these scores for the 500 samples with the lowest similarity in the first and last layers.
}
\label{tab:sim_score}
\end{table}

\begin{table}[t!]
    \centering
    \tiny    
    \resizebox{\columnwidth}{!}{%
        \begin{tabular}{l|cccc}         
        Top-k percent (\%)             & 30    & 20    & 15    & 70  \\
        \hline
        $\text{DeiT-Ti}_{r\text{=13}}$ & 65.16 & 52.83 & 43.29 & 91.80 \\
        $\text{DeiT-S}_{r\text{=13}}$  & 60.10 & 48.28 & 39.35 & 88.70 \\
        \hline
        \end{tabular}%
    }

\caption{
In the previous token merging method~\cite{bolya2023token}, we observed the overlap between the attentive tokens and the merged tokens in the first layer.
}
\label{tab:merged_topk}
\end{table}

\paragraph{Similarity Merging in Re-tokenization.}
\label{subsec:similarity_merging}
Previous token merging methods, e.g., ToMe~\cite{bolya2023token}, merge pairs of tokens with the highest similarity, selecting the fixed top-k most similar pairs per layer.
However, as merging proceeds progressively, tokens with low similarity are apt to be merged, resulting in semantically essential tokens to be diluted.
Tab.~\ref{tab:sim_score} shows a marked decline in the similarity score of ToMe from the first to the last layer during the merging procedures, indicating information loss.
This loss correlates with accuracy drops of 0.47\% in DeiT-S, as shown in Tab.~\ref{tab:main_result_merge}, despite not discarding any tokens in ToMe.

We also found that highly attentive tokens were primarily merged, as shown in Tab.~\ref{tab:merged_topk}.
For instance, in $\text{DeiT-Ti}_{r\text{=13}}$, 65.16\% of merged tokens are within the top-30\% attentive tokens in the first layer.
This suggests that important tokens, as indicated by their class attention scores, are predominantly merged, leading to potential information loss as the attentive tokens are quickly merged in the early layers. As the process progresses to later layers, the proportion of inattentive tokens increases, which may result in lower similarity during merging.

On the other hand, ImagePiece retokenizes inattentive tokens into semantically meaningful units, as shown in Tab.~\ref{tab:sim_score}, where the similarity scores of ImagePiece remain consistent or improve by 0.0005 from the first to the last layer.
Also, the similarity score of ImagePiece in the first layer is higher than that of ToMe, enabling a more effective merging process.
Unlike previous token merging methods, ImagePiece discards final non-semantic tokens based on the re-evaluation after retokenization, which benefits subsequent retokenizations by eliminating truly non-semantic tokens.

During the retokenization, we identified that the similarity between inattentive tokens in the early layer was not sufficiently high, complicating the decision on which token pairs to merge, so we enhance the semantics of these inattentive tokens for effective retokenization. 
To improve the semantics of tokens within a visual scene through local coherence, we utilized the local coherence bias module for the retokenization to aggregate locally coherent patches.
This approach tokenizes the image into tokens with overlapping patches, thereby enhancing token locality.
As shown in Tab~\ref{tab:loren_vs_patchify}, ours improves the locality of tokens, which, in turn, increases the average similarity score between inattentive tokens in the first layer.

\begin{table}[t]
    \centering
    \tiny
    \resizebox{0.8\columnwidth}{!}{%
    \begin{tabular}{l|c}
        Image tokenization method & Similarity score \\
        \hline
        Patch tokenizer of ViT & 0.5293 \\
        \rowcolor{lightblue}
        Local coherence bias (Ours) & 0.8091 \\
        \hline
    \end{tabular}
    }
    \caption{Comparison between the similarity scores obtained using patch tokenizer of ViTs and ours in the first layer.
    }
    \label{tab:loren_vs_patchify}
\end{table}

\renewcommand{\arraystretch}{1.25} % 1.5 times the default row height
\begin{table}[t]
    \centering
    \resizebox{\columnwidth}{!}{%
        \begin{tabular}{l|cc}
        \multirow{2}{*}{Model} & \multirow{2}{*}{Acc (\%)}  & Throughput \\
        &  & (img/s)  \\
        \hline
        DynamicViT~\cite{rao2021dynamicvit} & 79.32 & 3762.0 \\
        \rowcolor{lightblue}
        ImagePiece + DynamicViT & \textbf{80.11} \color{red}(+0.79) & 3857.1 \\
        \hline
        ToMe~\cite{bolya2023token} & 79.36 & 3806.1 \\
        \rowcolor{lightblue}
        ImagePiece + ToMe & \textbf{80.08} \color{red}(+0.72) & 3876.5 \\
        \hline
        EViT~\cite{liang2022evit} & 79.37 & 3787.5 \\
        \rowcolor{lightblue}
        ImagePiece + EViT & \textbf{80.22} \color{red}(+0.85) & 3891.9 \\
        \hline
        \end{tabular}%
    }
    \caption{Comparison of prior token reduction methods with and without the integration of ImagePiece on ImageNet-1k.
    By applying ImagePiece for retokenization, the performance of prior methods improves by an average of 0.79\%.}
    \label{tab:imagepiece_exp}
\end{table}

\paragraph{Compatibility of ImagePiece.}
\label{subsec:compatibility_of_ImagePiece}
We conducted an experiment to assess the impact of integrating our method, ImagePiece, into existing token reduction methods.
Specifically, we compared the performance of the existing methods with and without ImagePiece as a tokenizer.
% Tab.~\ref{tab:imagepiece_exp} in the supplement shows that our approach effectively enhances the efficiency of ViTs compared to traditional token reduction methods alone.

Following Tab.~\ref{tab:main_result}, we apply ImagePiece to two pruning strategies: pruning by learned projection layer~\cite{rao2021dynamicvit} and pruning by [CLS] attentiveness~\cite{liang2022evit}.
These pruning-based models, such as DynamicViT~\cite{rao2021dynamicvit} and EViT~\cite{liang2022evit}, integrate ImagePiece into the transformer blocks, excluding the specific pruning layers, to retokenize the tokens before reevaluating them for pruning.
% For instance, in DynamicViT, when tokens in the 3rd, 6th, 9th transformer blocks are discarded, ImagePiece is applied to all other layers.
% In the merging-based method, e.g., ToMe~\cite{bolya2023token}, ImagePiece is utilized similarly to the pruning-based methods.
% As non-semantic tokens can often be transformed into semantically meaningful units through retokenization, this helps ToMe to merge the tokens more accurately and improves performance.
In the merging-based method ToMe~\cite{bolya2023token} from Tab.~\ref{tab:main_result_merge}, ImagePiece retokenizes non-semantic tokens into meaningful units, enabling more accurate token merging and improved performance.

\paragraph{Ablation Study.}
To introduce local inductive bias into visual scenes, we incorporate a \textit{local coherence bias} module before the ImagePiece pipeline. 
The performance of our method without the \textit{local coherence bias} module is 79.81\%, demonstrating that the full ImagePiece framework with the \textit{local coherence bias} is effective, as our method achieves an accuracy of 80.22\%.

\section{Conclusion}
\label{sec:conc}

In this paper, we propose ImagePiece, a novel re-tokenization strategy for enhancing the efficiency of Vision Transformers.
ImagePiece merges non-semantic but locally coherent tokens into meaningful chunks (or discard them if they remain irrelevant to the visual scene), significantly increasing inference speed and improving ImageNet classification accuracy.
Our approach not only reduces the overall token count, but also ensures that the remaining tokens contribute more meaningfully to the overall visual understanding.
Extensive experiments and analysis demonstrate that our re-tokenization approach is more effective than previous patch tokenizers used in ViTs.
As for future work, this advancement sets a new standard in token management, promising significant impacts on efficient ViTs across various applications.

\end{document}